\documentclass[letterpaper]{article} 
\usepackage[preprint]{aaai2027}  
\usepackage[hyphens]{url}  
\usepackage{graphicx} 
\usepackage{natbib}  
\usepackage{caption} 
\usepackage{algorithm}
\usepackage{algorithmic}

\usepackage{newfloat}
\usepackage{listings}
\DeclareCaptionStyle{ruled}{labelfont=normalfont,labelsep=colon,strut=off} 
\floatstyle{ruled}
\newfloat{listing}{tb}{lst}{}
\floatname{listing}{Listing}

\usepackage{booktabs}

\title{SemRAG: Semantic-Guided Sentence-Level Retrieval-Augmented Sign Language Production}

\author{
    Xiao Liu\equalcontrib,
    Shiwei Gan\equalcontrib,
    Yafeng Yin\corresponding,
    Peishuo Liu,
    Bowen Guo,
    Kuizhuang Liu,\\
    Cong Wang,
    Lei Xie
}

\affiliations{
    State Key Laboratory of Novel Software Technology, Nanjing University\\
    liuxiaox@smail.nju.edu.cn,
    \{sw,yafeng\}@nju.edu.cn,\\
    \{peishuo,bowen,kuizhuang\}@smail.nju.edu.cn,
    cw@nju.edu.cn, lxie@nju.edu.cn
}

\usepackage{graphicx} 
\usepackage{booktabs}
\usepackage{multibib}
\usepackage{pifont}

\usepackage{caption}
\usepackage{subcaption}
\usepackage{multirow}  
\usepackage{amsthm,amsmath,amssymb}
 
\usepackage{adjustbox}
\usepackage{mathrsfs}
\usepackage{bbding}
\usepackage{url}            %
\usepackage[table]{xcolor} 

\definecolor{LightBlue}{RGB}{240,248,255}
\definecolor{deepgray}{gray}{0.3}

\newlength\savewidth

\makeatletter\renewcommand\paragraph{\@startsection{paragraph}{4}{\z@}
  {.5em \@plus1ex \@minus.2ex}{-.5em}{\normalfont\normalsize\bfseries}}\makeatother

\newcolumntype{x}[1]{>{\centering\arraybackslash}p{#1pt}}
\newcolumntype{y}[1]{>{\raggedright\arraybackslash}p{#1pt}}
\newcolumntype{z}[1]{>{\raggedleft\arraybackslash}p{#1pt}}

\newcommand{\app}{\raise.17ex\hbox{$\scriptstyle\sim$}}

\definecolor{baselinecolor}{gray}{.9}

\newcolumntype{*}{>{\global\let\currentrowstyle\relax}}
\newcolumntype{^}{>{\currentrowstyle}}

\definecolor{dt}{gray}{0.7}  %

\usepackage[capitalize]{cleveref}
\crefname{section}{Sec.}{Secs.}
\Crefname{section}{Section}{Sections}
\Crefname{table}{Table}{Tables}
\crefname{table}{Tab.}{Tabs.}

\newcolumntype{S}{@{}>{\lrbox0}l<{\endlrbox}}  %
\definecolor{lightgreen}{HTML}{D8ECD1}

\newcommand{\eg}{{\emph{e.g.}}, }

\usepackage{tikz}

\begin{document}

\maketitle

\begin{abstract}
Sign language production (SLP) aims to generate sign pose sequences that faithfully express spoken-language text while preserving the rhythm and coordination patterns of signing. 
Common SLP methods learn text-to-pose mappings from training data but often suffer from semantic inaccuracies in long or complex sentences.
Retrieval augmentation offers a way to introduce realistic signing priors to provide more explicit guidance for text-to-pose generation.
However, existing retrieval-based SLP methods typically operate at the gloss or word level, relying on gloss-pose dictionaries or isolated sign segments that require costly annotations and fail to preserve continuous sentence-level dynamics.
To bridge this gap, we propose SemRAG, a semantic-guided sentence-level retrieval-augmented framework that retrieves full-sentence signing templates as priors without constructing gloss/word-level sign dictionaries. 
While such templates preserve local motion patterns and global rhythm, they also introduce a semantic mismatch problem: irrelevant motions in the retrieved sentence may interfere with the target semantics.
To address this issue, we introduce an SLT-guided template semantic localization module that identifies target-relevant temporal regions within the retrieved template while suppressing irrelevant signing content. 
To adaptively incorporate the template features into the generator, we further introduce layer-specific learnable gates that regulate their contribution across generator layers.
Extensive experiments show that SemRAG improves semantic consistency and motion quality over baselines under both back-translation and motion accuracy metrics.

\end{abstract}

\section{Introduction} 
\label{sec:introduction}
Sign language video generation (SLVG)~\cite{saunders2022signing,fang2023signdiff} aims to synthesize signing videos from spoken-language text by jointly modeling linguistic semantics, body dynamics, and visual appearance. However, directly synthesizing both signing motion and visual appearance makes temporally consistent video generation particularly challenging. As a structured, pose-based formulation, sign language production (SLP)~\cite{saunders2020adversarial,rastgoo2021sign,stoll2022there} maps text into pose sequences, providing an explicit representation of signing behavior. A desirable SLP system should preserve input semantics while generating natural and temporally coherent signing sequences.

Existing text-to-pose models typically encode the input text to provide semantic conditions and then either directly regress pose sequences in the original pose space~\cite{saunders2020progressive,tang2025gloss} or generate pose representations in a learned latent space using generative models, such as diffusion models~\cite{xie2024g2p,liu2026signpr} and Transformer-based decoders~\cite{yin2024t2s,zuo2025signs,ma2024ms2sl}.
Although these models generate plausible signing poses, they rely on sequence-level supervision and lack explicit grounding between words or phrases and their corresponding pose segments. Consequently, generated sequences may contain inaccurate motions, incorrect temporal ordering, or incomplete semantic expression, particularly for long or complex sentences.

\begin{figure}[t]
  \centering
    \includegraphics[width=1.0\columnwidth]{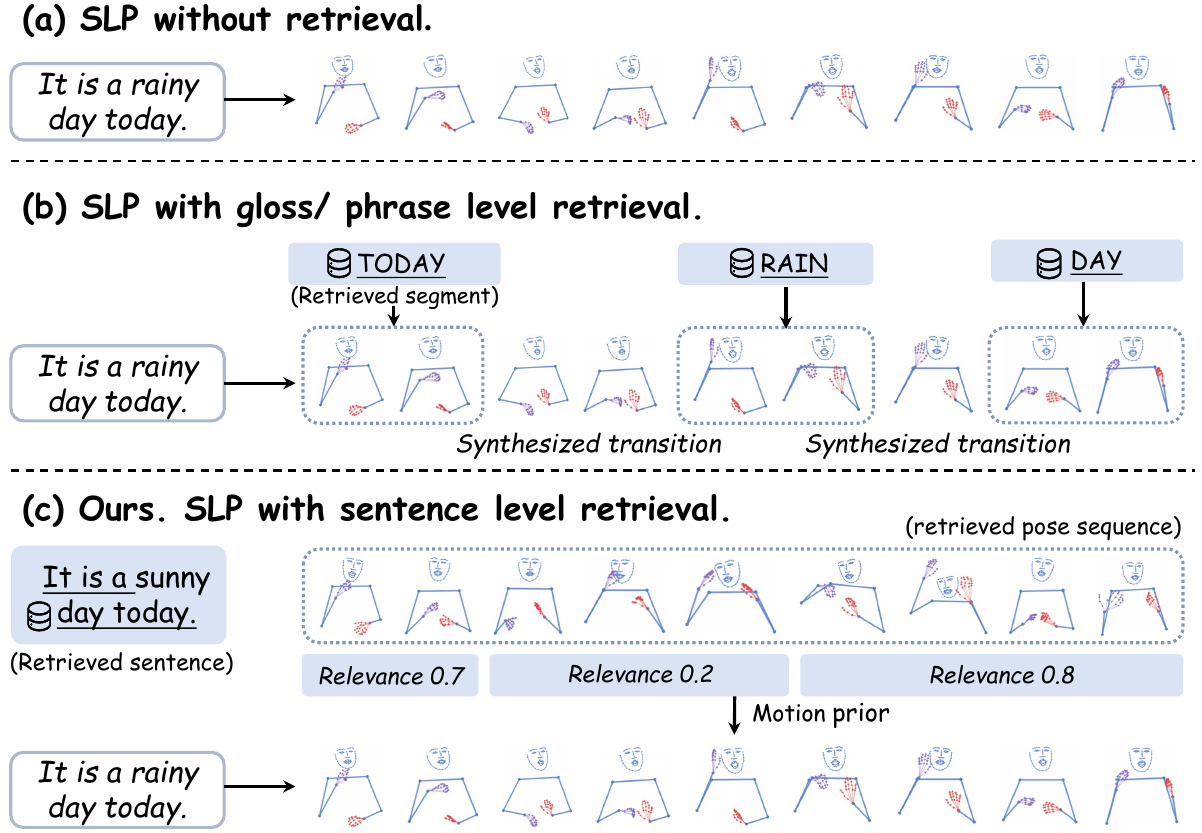}
    \vspace{-2mm}
   \caption{Previous SLP models vs our sentence-level retrieval-augmented SLP method.}
   \label{fig:motivation} 
  \vspace{-5mm}
\end{figure} 

One promising direction to alleviate semantic drift is retrieval-augmented generation~\cite{zhang2023remodiffuse,yu2025remogpt,xu2026vimorag}. 
Existing retrieval-based SLP methods often rely on gloss-level~\cite{saunders2022signing,zuo2024simple,walsh2024sign} or word-level~\cite{zuo2025signs} motion priors. 
They typically construct a gloss/word-pose dictionary, where each gloss or spoken-language word is associated with a pose segment obtained from continuous signing videos through automatic or manual temporal localization, such as continuous sign language recognition (CSLR)-based segmentation or expert annotations, or collected from external sign dictionaries~\cite{zuo2024simple}. 
Given an input text, these methods first predict a gloss sequence or directly match spoken-language words, retrieve the corresponding pose segments from the dictionary, and then concatenate and smooth them to produce the final pose sequence. 
Despite their explicit gloss/word-to-pose correspondence, such methods have two main limitations. 
First, existing approaches either depend on expensive human annotations (\eg gloss labels or word-level sign dictionaries), or rely on automatically extracted segments, whose localization errors may introduce irrelevant motion priors.
Second, isolated gloss- or word-level segments discard their original sentence context. Although transition modules can smooth adjacent segments, sentence-level timing and long-range coordination must still be reconstructed from independently retrieved units.
Sequence-level retrieval has been explored in motion and video generation to provide reusable priors~\cite{zhang2023remodiffuse,xu2026vimorag,yu2025remogpt}.  
\textbf{These observations motivate a question: can retrieval-augmented SLP similarly benefit from sentence-level signing templates?}

Retrieving full-sentence signing templates paired with semantically similar texts offers a promising solution. 
As illustrated in Figure~\ref{fig:motivation}, unlike retrieval based on isolated gloss- or word-level segments, sentence-level templates preserve continuous signing dynamics, including temporal organization, movement rhythm, and cross-sign transitions.
Similar phrases may provide local motion cues, while similar sentence structures may offer broader guidance for the overall signing progression. However, directly transferring existing sequence-level retrieval methods from motion and video generation to SLP overlooks the dense linguistic content of sign pose sequences, where even small textual differences may correspond to critical local signing motions. \textbf{This creates a critical mismatch problem: the retrieved template may contain irrelevant motion fragments that are semantically inconsistent with the target sentence.}

To address this problem, we propose SemRAG, a semantic-guided sentence-level retrieval-augmented framework for sign language production.
Given an input sentence, SemRAG retrieves a semantically related sentence–pose pair from a training-only template set and uses its signing sequence as an informative but imperfect reference.
Rather than directly copying or uniformly injecting the retrieved sequence, SemRAG selectively exploits input-relevant temporal regions while suppressing semantically mismatched motions. 
Specifically, we use decoder cross-attention from a frozen SLT model to estimate which temporal pose regions correspond to tokens in the retrieved sentence. 
We then aggregate the attention responses of content words shared by the input and retrieved sentences into a temporal relevance gate that reweights the template features.
The resulting template features are incorporated through layer-specific learnable gates, allowing individual generator layers to adaptively regulate the contribution of the retrieved reference while preserving target-specific semantics.
The main contributions of this work are summarized as follows:

\begin{itemize}
    \item To move beyond isolated gloss/word-level motion priors, we propose SemRAG, a sentence-level retrieval-augmented SLP framework that retrieves semantically related text--pose pairs and uses their full signing sequences as references to guide text-to-pose generation.
    
    \item To mitigate semantic interference from imperfect retrieved templates, we propose an SLT-guided semantic localization module to identify target-relevant temporal regions and a layer-gated mechanism to regulate template contributions across generator layers.

    \item Extensive experiments conducted on Phoenix14T and CSL-Daily demonstrate the effectiveness of SemRAG in improving semantic consistency and motion quality compared with text-to-pose and retrieval-based baselines.
\end{itemize}

\section{Related Work}

\begin{figure*}[t]
  \centering
    \includegraphics[width=2.0\columnwidth]{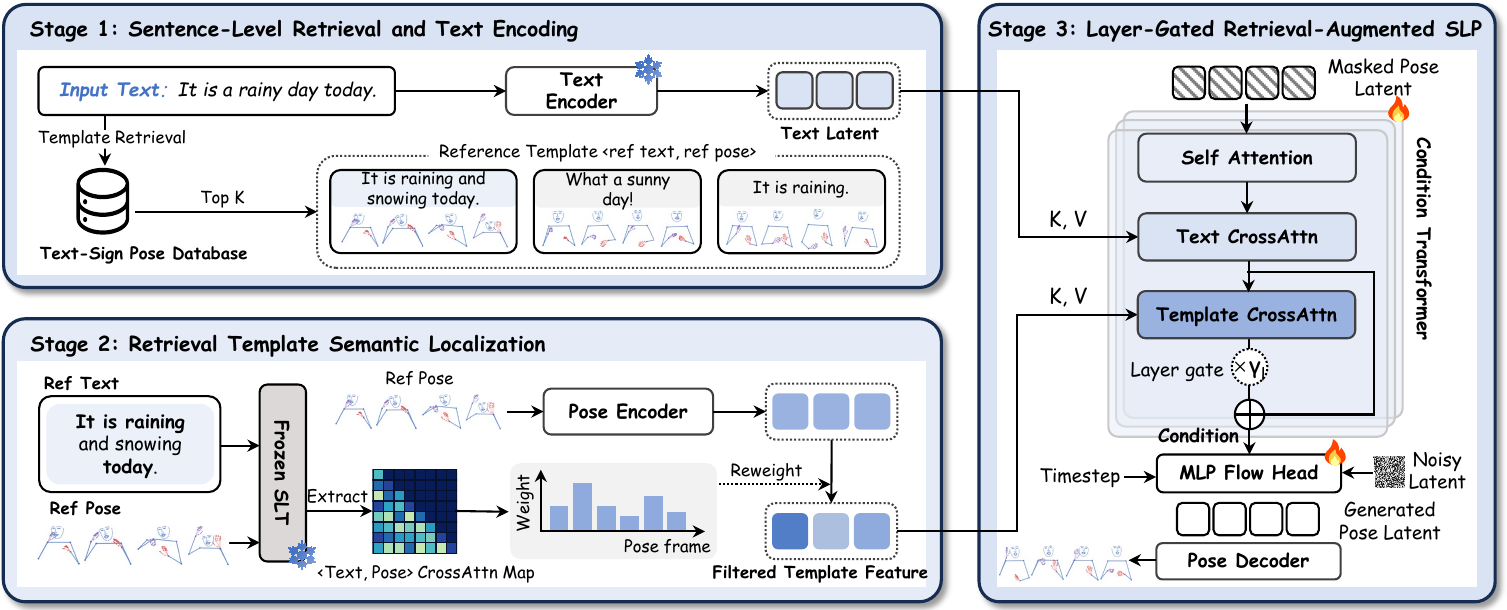}
    \vspace{-1mm}
   \caption{Overview of SemRAG.
Given an input sentence, SemRAG retrieves sentence-level signing templates and encodes text features (Stage 1), localizes semantically relevant template regions using a frozen SLT model and builds filtered template features (Stage 2), and performs layer-gated template injection into a generator for SLP (Stage 3).}
   \label{fig:framework} 
  \vspace{-2mm}
\end{figure*}

\paragraph{Sign Language Production.} 
Sign language production (SLP)~\cite{saunders2020adversarial, tang2025gloss}~\nocite{bensabath2026text,walsh2024data,saunders2022signing,saunders2021mixed,stoll2018sign,stoll2022there,saunders2020progressive,baltatzis2024neural,ma2024ms2sl,wang2025linguistics,bensabath2026text,walsh2024data,saunders2022signing,saunders2021mixed,stoll2018sign,stoll2022there,saunders2020progressive,baltatzis2024neural,ma2024ms2sl,wang2025linguistics,baltatzis2024neural,camgoz2018neural,camgoz2020sign,chen2022two,fang2023signdiff,fardo2016formal,gong2024llms,gu2022vector,guo2023animatediff,heusel2017gans, hu2024animate,lin2004automatic,ma2024ms2sl, qi2024signgen, rastgoo2021sign, razavi2019generating, saunders2020adversarial, saunders2020progressive, saunders2021continuous, saunders2021mixed, saunders2022signing, shi2022open, shi2024pose, stoll2018sign, stoll2022there, tang2025gloss, villegas2022phenaki, walsh2024data, wang2025advanced, wei2023improving, xie2024g2p, yin2023gloss, zuo2025signs, unterthiner2018towards,qi2024signgen} aims to convert spoken-language text into sign language sequences that faithfully convey the input semantics while maintaining natural temporal dynamics. 
Based on whether external motion priors are incorporated, existing SLP methods can be broadly categorized into non-retrieval and retrieval-based approaches. Non-retrieval methods generate sign language sequences, such as pose sequences~\cite{yu2026focal} or SMPL-X mesh sequences~\cite{yin2024t2s}, solely from text or gloss conditions. Early studies~\cite{saunders2021continuous} adopt sequence-to-sequence architectures to regress sign poses from text inputs, while recent works~\cite{xie2024g2p} further introduce learned latent pose representations and autoregressive or diffusion-based generation frameworks to improve the generated poses quality. These methods condition the generation process on textual or gloss representations encoded by pretrained language models or task-specific encoders. However, such linguistic conditions are typically learned through sequence-level supervision or only weakly aligned with pose tokens, providing limited explicit grounding between local semantic units and their corresponding motion segments. Consequently, semantic misalignment or drift may arise, particularly for long sentences or syntactically complex expressions.

Retrieval-based methods seek to alleviate this issue by introducing real sign motion priors, such as gloss/word-pose dictionaries~\cite{saunders2022signing,zuo2024simple,walsh2024sign} and isolated sign segments~\cite{zuo2025signs}. 
These methods typically associate glosses or spoken-language words with pose segments obtained from isolated sign dictionaries or temporally segmented signing videos. 
Given an input text or predicted gloss sequence, they retrieve corresponding pose segments from the retrieval set and concatenate, smooth, or refine them with additional generation modules to produce the signing sequence. 
Representative methods, such as SOKE~\cite{zuo2025signs} and Sign Stitching~\cite{walsh2024sign}, demonstrate the usefulness of retrieved sign segments as explicit local motion references. 
However, this paradigm often requires costly gloss/word annotations, external sign dictionaries, or reliable CSLR-based temporal segmentation. 
Moreover, concatenating isolated gloss- or word-level segments may disrupt transition dynamics and sentence-level rhythm in continuous signing. 
To address these limitations, we explore sentence-level retrieval and develop an SLT-guided semantic filtering and layer-gated injection mechanism that selectively incorporates useful motion cues from retrieved templates into the text-to-pose generation process.

\paragraph{Retrieval-Augmented Generation.}
Retrieval-augmented generation introduces examples or knowledge into the generation process, allowing a model to condition not only on the input but on retrieved references from a database~\cite{borgeaud2022improving}. Although studied in natural language processing~\cite{lewis2020retrieval}, this paradigm has been extended to visual and motion generation, where retrieved images, videos, or motion clips provide appearance and motion priors.
Existing video and motion retrieval methods~\cite{zhang2023remodiffuse,xu2026vimorag} primarily transfer reusable visual or motion patterns, whereas SLP requires fine-grained linguistic correspondence, since even small textual differences may imply critical local signing changes.
In SLP, prior works such as SOKE~\cite{zuo2025signs} and Sign Stitching~\cite{walsh2024sign} also exploit retrieved pose references for generation. However, these methods mainly retrieve glosses, isolated signs, or local motion segments, while full-sentence signing sequences remain largely unexplored as retrieval references. Sentence-level templates can preserve global rhythm and body-motion patterns, but even small textual differences may cause them to contain local signing motions inconsistent with the target sentence. 
In contrast, our work investigates sentence-level retrieval-augmented SLP and exploits useful temporal regions of retrieved signing templates while effectively suppressing semantically inconsistent content.

\section{Method}

\paragraph{Overview.}
Given an input text sentence \(x=\{w_m\}_{m=1}^{M}\), the goal of sign language production is to generate a sign pose sequence 
\(\hat{P}=\{\hat{p}_s\}_{s=1}^{S}\), where \(\hat{p}_s\in\mathbb{R}^{J\times C}\) denotes \(J\) body keypoints with \(C\)-dimensional coordinates at the \(s\)-th frame, and \(S\) is the number of generated frames. 
As illustrated in Figure~\ref{fig:framework}, we propose SemRAG, a semantic-guided sentence-level retrieval-augmented framework for text-to-pose sign language production. 
The overall process can be summarized as follows.
First, given the input sentence \(x\), SemRAG retrieves a semantically related sentence-pose pair \((x_r,P_r)\) from a training-only template set, where \(P_r=\{p_t^r\}_{t=1}^{S_r}\) is the retrieved signing sequence and \(S_r\) denotes its frame length. 
The retrieved sequence is encoded into latent template features \(H_r\), which provide real sign pose priors. 
Second, since \(P_r\) may contain motions irrelevant to \(x\), we use a frozen SLT model to derive a soft temporal gate \(g\) from decoder cross-attention maps. 
The gate reweights the retrieved template features, producing the SLT-filtered template features \(\hat{H}_r=H_r \odot \mathbf{g}\). 
Third, in the latent motion space defined by the temporal pose VAE, a layer-gated retrieval-augmented generator predicts the target latent sequence \(\hat{Z}=\{\hat{z}_t\}_{t=1}^{T_z}\), conditioned on token-level T5 text representations \(H_x=f_{\mathrm{text}}(x)\) and the filtered template features \(\hat{H}_r\).
Finally, the frozen VAE decoder \(\mathcal{D}\) reconstructs the generated latent sequence into the sign pose sequence, \(\hat{P}=\mathcal{D}(\hat{Z})\).

\subsection{Sentence-Level Template Retrieval and Semantic Localization}
To preserve continuous signing dynamics without relying on gloss/word-pose dictionaries or isolated pose segments, SemRAG retrieves complete signing sequences as sentence-level templates. However, a retrieved template may only partially match the input semantics and may contain irrelevant motions. We therefore perform sentence-level retrieval followed by SLT-guided semantic localization to produce filtered template features that serve as selective motion priors during subsequent text-to-pose generation.

\paragraph{Sentence-level Template Retrieval.}
Given the training set of paired sentences and sign pose sequences
\(\mathcal{D}_{\mathrm{tr}}=\{(x_i,P_i)\}_{i=1}^{N}\),
we construct a sentence-level template set from the training split.
Each entry consists of a sentence and its corresponding sign pose sequence.
For an input sentence \(x\), we retrieve the top-\(K\) relevant sentence--pose pairs from this fixed set.
During training, the paired sample of \(x\) is excluded from the candidate set to avoid self-retrieval.

To retrieve templates that are both semantically related and lexically consistent, we adopt a hybrid retrieval strategy.
For each candidate sentence \(x_i\), we compute a hybrid retrieval score by combining sentence-level semantic similarity, content-token overlap, and bigram overlap:
\[
s(x,x_i)=
\lambda_{\mathrm{sem}}s_{\mathrm{sem}}(x,x_i)
+\lambda_{\mathrm{tok}}s_{\mathrm{tok}}(x,x_i)
+\lambda_{\mathrm{bi}}s_{\mathrm{bi}}(x,x_i).
\]
Sentence-level similarity captures global semantics but may overlook lexical differences linked to local signing motions. We therefore add content-token overlap for semantic-unit coverage and bigram overlap for local phrase structure and word order.
Here, \(s_{\mathrm{sem}}\) is the cosine similarity between sentence embeddings encoded by T5 text encoder, \(s_{\mathrm{tok}}\) measures the overlap between normalized content-token sets, and \(s_{\mathrm{bi}}\) measures the overlap between normalized token-bigram sets. 
The weights \(\lambda_{\mathrm{sem}}, \lambda_{\mathrm{tok}}, \lambda_{\mathrm{bi}}\) control the relative contributions of the three retrieval cues and are fixed for all experiments. 
Detailed definitions of these scores and the weight settings are provided in the appendix.
We rank all candidates by \(s(x,x_i)\) and retain the top-\(K\) sentence-pose pairs:
\[
\mathcal{R}_K(x)=\{(x_r^k,P_r^k)\}_{k=1}^{K}.
\]
If the best retrieval score is below a predefined threshold, we set \(H_r=\mathbf{0}\) and disable template conditioning. 
Otherwise, the retrieved pose sequence \(P_r\) is encoded by the frozen pose tokenizer into latent template features \(H_r\), which are passed to the semantic localization module.

\paragraph{SLT-Guided Temporal Semantic Localization.}
To retain useful template cues while suppressing semantically mismatched regions, we use a frozen pretrained pose-based SLT model to estimate temporal relevance over the retrieved signing sequence.
The decoder cross-attention maps of the SLT model provide weak cues for localizing temporal regions associated with text tokens.
We aggregate the attention responses of content words shared by the input and retrieved sentences and convert them into a soft temporal gate, which reweights the retrieved template features before injection.

Formally, let \(x\) be the input sentence, and let \((x_r,P_r)\) denote the retrieved sentence--pose pair.
We feed \((x_r,P_r)\) into the frozen SLT model under teacher forcing and extract its decoder-to-encoder cross-attention maps.
For decoder layer \(l\) and head \(h\), we denote the attention map by \(A^{l,h}\), where each row corresponds to a text token in \(x_r\) and each column corresponds to a pose token encoded from \(P_r\). We define \(\mathcal{W}\) as the indices of content tokens in \(x_r\) that also occur in the input sentence \(x\), after normalization and stop-word removal.
We use the attention map from the selected decoder layer \(l^\ast\) and average the decoder cross-attention of the matched content tokens over all heads:
\[
\tilde{\mathbf{g}}
=
\frac{1}{|\mathcal{H}|\,|\mathcal{W}|}
\sum_{h\in\mathcal{H}}
\sum_{w\in\mathcal{W}}
A^{l^\ast,h}_{w,:}.
\]
We then apply min--max normalization to \(\tilde{\mathbf{g}}\) to obtain a soft temporal gate \(\mathbf{g}\) over the retrieved pose sequence. A larger value indicates that the corresponding pose token is more strongly associated with the matched content tokens according to the frozen SLT model.
Finally, since the SLT pose tokens and pose latents may have different temporal resolutions, we resize \(\mathbf{g}\) to the latent length of the retrieved template and use it to reweight the template features:
\[
\hat{H}_r
=
H_r \odot \mathbf{g},
\]
where \(H_r\in\mathbb{R}^{T_z^r\times d}\) denotes the template features, \(T_z^r\) is the latent length of the retrieved template, and the resized gate is broadcast along the feature dimension.
The resulting \(\hat{H}_r\) provides filtered template features for the latent generator; 
for $K>1$, the independently filtered features of all retrieved templates are concatenated along the temporal dimension.

\subsection{Layer-Gated Retrieval-Augmented Generation}
After obtaining the filtered template features \(\hat{H}_r\), SemRAG generates the target pose latents with a retrieval-augmented latent flow-matching generator.
The input text provides the primary semantic condition, while the retrieved template serves as an auxiliary motion prior.
Template information is injected through learnable layer gates rather than with equal strength across all generator layers.

\paragraph{Latent Flow Matching Generator.}
We generate sign poses in the temporally compact latent space of a pose tokenizer that follows the temporal pose VAE architecture used in MotionStreamer~\cite{xiao2025motionstreamer}. Its encoder \(\mathcal{E}\) maps a pose sequence \(P\) into latent tokens
\[
Z=\mathcal{E}(P)=\{z_t\}_{t=1}^{T_z},
\]
and whose decoder reconstructs poses as \(\tilde{P}=\mathcal{D}(Z)\).
The tokenizer is pretrained and frozen during generator training.

In this latent space, the generator consists of two components: a condition Transformer that models contextual dependencies among partially masked pose latents, text features, and retrieved template features, and a flow-matching head that predicts conditional velocity fields at the masked positions.
During training, we randomly select a subset of latent positions and replace their tokens with a learnable mask embedding, yielding a masked context sequence \(\tilde{Z}\).
The unmasked tokens provide motion context, while the masked positions indicate the latents to be generated.
The input sentence is encoded by a pretrained T5 encoder into token-level text representations \(H_x\).
The condition Transformer processes \(\tilde{Z}\) using latent self-attention, text cross-attention over \(H_x\), and layer-gated template cross-attention over \(\hat{H}_r\), producing contextual features
\(
C=\{c_t\}_{t=1}^{T_z}.
\)
These features condition the flow-matching prediction at the masked positions.
For each masked latent position \(t\), we sample Gaussian noise
\(z_t^0\sim\mathcal{N}(0,I)\) and a flow time
\(\tau\sim\mathcal{U}(0,1)\), and construct
\[
z_t^\tau=\tau z_t+(1-\tau)z_t^0.
\]
The flow-matching head takes \(z_t^\tau\), \(\tau\), and the corresponding contextual feature \(c_t\) as input and predicts the conditional velocity
\(
v_\theta(z_t^\tau,\tau,c_t).
\)
The flow-matching objective is
\[
\mathcal{L}_{\mathrm{flow}}
=
\mathbb{E}
\left[
\left\|
v_\theta(z_t^\tau,\tau,c_t)
-
\left(z_t-z_t^0\right)
\right\|_2^2
\right],
\]
where the expectation is taken over training samples, masked latent positions, noise samples, and flow times.

During inference, given a target latent length \(T_z\), all positions are initialized with the learnable mask embedding.
At each iteration, the condition Transformer computes contextual features from the partially generated latent sequence, the T5 text features \(H_x\), and the filtered template features \(\hat{H}_r\).
For selected masked positions, the flow-matching head generates latent tokens by integrating the predicted velocity field, progressively reducing the number of masked positions.
Once all positions are filled, the completed latent sequence is decoded by \(\mathcal{D}\) into the final pose sequence \(\hat{P}\).

\paragraph{Layer-Gated Template Injection.}
Although filtered by SLT-guided localization, the retrieved template remains an auxiliary motion prior rather than a target sequence to be copied.
Uniformly injecting template information across condition Transformer layers ignores their different roles in representation learning and may weaken text-conditioned generation.
We therefore introduce learnable layer gates to control retrieval strength across Transformer layers.

Let \(H_l\) denote hidden states after text cross-attention in the \(l\)-th condition Transformer layer.
We compute a template residual by attending to the filtered template features:
\[
R_l=\operatorname{CrossAttn}(H_l,\hat{H}_r,\hat{H}_r),
\]
where \(H_l\) serves as the query, while \(\hat{H}_r\) provides the corresponding keys and values.
The resulting residual is then injected through a layer-specific learnable scalar gate:
\[
H_l' = H_l + \gamma_l R_l,
\qquad
\gamma_l=\sigma(\alpha_l),
\]
where \(\alpha_l\) is a learnable scalar for the \(l\)-th layer and
\(\sigma(\cdot)\) denotes the sigmoid function.
The layer gates are jointly optimized under the flow-matching objective to determine the contribution of retrieved template cues at different depths.

\subsection{Training and Inference}
SemRAG is trained in two stages.
We first train the sign pose VAE as the pose tokenizer using reconstruction, velocity, and KL losses.
The trained encoder \(\mathcal{E}\) and decoder \(\mathcal{D}\) are then frozen.
In the second stage, we train the retrieval-augmented latent flow generator while keeping both the pose tokenizer and the pretrained SLT model frozen.
For each training sample, its corresponding sentence--pose pair is excluded from the template set to prevent self-retrieval.
The generator is optimized using the flow-matching objective.

During inference, templates are retrieved exclusively from the training template set.
If the highest retrieval score falls below a predefined threshold, template conditioning is disabled and SemRAG falls back to text-only generation.
The generated latent sequence is finally decoded by the frozen tokenizer decoder \(\mathcal{D}\) into the sign pose sequence \(\hat{P}\).

\section{Experiments}
\begin{table*}[t]
\centering

\scriptsize
\setlength{\tabcolsep}{2.5pt}
\renewcommand{\arraystretch}{1.05}

\resizebox{\linewidth}{!}{%
\begin{tabular}{lcccccccccccccccc}
\toprule

\multirow[c]{3}{*}{Method}
& \multicolumn{8}{c}{Phoenix14T}
& \multicolumn{8}{c}{CSL-Daily} \\
\cmidrule(lr){2-9}
\cmidrule(lr){10-17}

& \multicolumn{4}{c}{Dev}
& \multicolumn{4}{c}{Test}
& \multicolumn{4}{c}{Dev}
& \multicolumn{4}{c}{Test} \\
\cmidrule(lr){2-5}
\cmidrule(lr){6-9}
\cmidrule(lr){10-13}
\cmidrule(lr){14-17}

& R$\uparrow$
& B4$\uparrow$
& MPJPE$\downarrow$
& Hand$\downarrow$
& R$\uparrow$
& B4$\uparrow$
& MPJPE$\downarrow$
& Hand$\downarrow$
& R$\uparrow$
& B4$\uparrow$
& MPJPE$\downarrow$
& Hand$\downarrow$
& R$\uparrow$
& B4$\uparrow$
& MPJPE$\downarrow$
& Hand$\downarrow$ \\
\midrule

\multicolumn{17}{l}{\textit{Without retrieval}} \\
\midrule

PT$^\ast$~\cite{saunders2020progressive}
& 11.87 & 3.88 & 40.63 & 10.47
& 13.17 & 4.31 & 50.80 & 10.26
& 7.21 & 0.39 & 87.92 & 13.97
& 7.54 & 0.41 & 90.28 & 13.29 \\

SignPR~\cite{liu2026signpr}
& 30.84 & 9.12 & 23.77 & 1.90
& 32.86 & 9.41 & 23.04 & 1.87
& 15.02 & 3.36 & 44.59 & 2.07
& 14.43 & 3.01 & 44.22 & 2.09 \\

MoLingo$^\ast$~\cite{he2026molingo}
& 31.43 & 9.57 & 25.28 & 1.88
& 32.06 & 9.83 & 24.85 & 1.95
& 16.33 & 1.88 & 23.26 & 2.06
& 15.97 & 1.72 & 23.03 & 2.11 \\

\midrule
\multicolumn{17}{l}{\textit{Retrieval-based methods}} \\
\midrule

ReMoDiffuse$^\ast$~\cite{zhang2023remodiffuse}
& 26.54 & 7.55 & 25.74 & 3.26
& 27.18 & 6.73 & 25.60 & 3.11
& 21.94 & 6.39 & 22.39 & 1.96
& 22.42 & 6.83 & 22.28 & 2.04 \\

SOKE$^\ast$~\cite{zuo2025signs}
& 30.94 & 10.15 & 23.65 & 1.91
& 31.23 & 10.43 & 22.94 & 1.85
& 27.85 & 9.74 & 22.41 & 2.05
& 28.29 & 10.31 & 22.50 & 2.17 \\

\midrule

\rowcolor{gray!10}
\textbf{SemRAG$^a$}
& \textbf{33.75}
& \textbf{11.00}
& \textbf{23.68}
& \textbf{1.69}
& \textbf{34.64}
& \textbf{11.42}
& \textbf{22.59}
& \textbf{1.71}
& \textbf{28.34}
& \textbf{11.50}
& \textbf{22.38}
& \textbf{1.92}
& \textbf{29.08}
& \textbf{12.15}
& \textbf{22.68}
& \textbf{2.00} \\

\rowcolor{gray!10}
\textbf{SemRAG$^b$}
& \textbf{34.22}
& \textbf{11.35}
& \textbf{22.70}
& \textbf{1.64}
& \textbf{35.09}
& \textbf{12.29}
& \textbf{22.47}
& \textbf{1.56}
& \textbf{28.41}
& \textbf{11.55}
& \textbf{21.56}
& \textbf{1.89}
& \textbf{29.21}
& \textbf{12.37}
& \textbf{22.75}
& \textbf{1.99} \\

\bottomrule
\end{tabular}
}
\vspace{1mm}

\parbox{\linewidth}{\scriptsize
$^\ast$ denotes reimplemented results.
R and B4 denote ROUGE-L and BLEU-4 back-translation scores, respectively, both computed using the same frozen MixSignGraph evaluator.
Hand denotes hand DTW-PA-JPE.
SemRAG$^a$ and SemRAG$^b$ use frozen MixSignGraph and TwoStream-SLT models as attention probes, respectively.
}
\vspace{-2mm}
\caption{SLP performance comparison on Phoenix14T and CSL-Daily.}
\label{tab:SLP}
\vspace{-1mm}
\end{table*}

\paragraph{Datasets.}
We evaluate SemRAG on Phoenix14T~\cite{camgoz2018neural} and CSL-Daily~\cite{zhou2021improving}. 
Phoenix14T contains 7,096/519/642 train/dev/test samples from 9 signers, with 1,066 glosses and 2,887 German words. 
CSL-Daily contains 18,401/1,077/1,176 samples from 10 signers, with 2,000 glosses and 2,343 Chinese words. 
We follow the standard splits of both datasets. 
Following SignPR~\cite{liu2026signpr}, we represent each signing sequence with 2D keypoints extracted by HRNet~\cite{sun2019deep}, including 42 hand, 68 face, and 11 upper-body keypoints.

\paragraph{Metrics.}
We evaluate semantic consistency and motion accuracy.
For semantic consistency, following SignPR~\cite{liu2026signpr}, we translate generated poses back to spoken language with a frozen SLT model~\cite{gan2025mixsigngraph} and report ROUGE-L~\cite{lin2004automatic} and BLEU-4~\cite{papineni2002bleu}.
For motion accuracy, we report MPJPE and hand DTW-PA-JPE.
MPJPE measures the average per-frame keypoint error, while hand DTW-PA-JPE first aligns generated and reference sequences with dynamic time warping and then computes Procrustes-aligned hand joint errors.

\paragraph{Implementation Details.}
SemRAG is trained in two stages.
We first train a temporal pose VAE following the architecture used in MotionStreamer~\cite{xiao2025motionstreamer} with AdamW for 300 epochs using a learning rate of \(5\times10^{-5}\) and batch size 4, and then freeze it as the pose tokenizer.
The VAE has a latent width of 16 and two stride-2 temporal downsampling stages.
For sentence-level retrieval, we set \(K=1\), use the hybrid retrieval score, and disable template conditioning when the top-ranked score is below 0.6.
The SLT-guided temporal gate is derived from decoder-to-encoder cross-attention maps at decoder layer 10 of a frozen SLT model.
The latent generator consists of a 16-layer condition Transformer with hidden dimension 1024 and 16 attention heads, followed by an 8-layer MLP flow head with hidden width 1280.
It is trained with AdamW for 500 epochs using a learning rate of \(1\times10^{-5}\) and batch size 4.
We use a cosine schedule for latent masking, classifier-free text dropout of 0.1, and retrieval-condition dropout of 0.1.
At inference, we adopt an oracle-length setting to isolate pose generation from length prediction errors and perform flow sampling with 32 steps.


\begin{table}[t]
\centering
\scriptsize
\setlength{\tabcolsep}{3.8pt}
\renewcommand{\arraystretch}{1.05}

\begin{tabular*}{0.85\columnwidth}{@{\extracolsep{\fill}}lcccc}
\toprule
Variant & R$\uparrow$ & B4$\uparrow$ & MPJPE$\downarrow$ & Hand$\downarrow$ \\
\midrule

\rowcolor{black!8}
\multicolumn{5}{l}{\textit{Component ablation}} \\
Text-only & 31.24 & 9.94 & 25.15 & 1.90 \\
+ retrieved template & 33.76 & 11.02 & 24.08 & 1.69 \\
+ SLT-filtered & 34.32 & 11.58 & 22.50 & 1.57 \\
+ layer gate & \textbf{35.09} & \textbf{12.29} & \textbf{22.47} & \textbf{1.56} \\

\addlinespace[3pt]
\rowcolor{black!8}
\multicolumn{5}{l}{\textit{Filtering mechanism analysis}} \\
No gate & 33.76 & 11.02 & 24.08 & 1.69 \\
Shuffled SLT gate & 26.28 & 6.59 & 25.27 & 1.84 \\
Inverted SLT gate & 24.23 & 5.68 & 25.29 & 1.83 \\
SLT gate & \textbf{35.09} & \textbf{12.29} & \textbf{22.47} & \textbf{1.56} \\

\bottomrule
\end{tabular*}
\vspace{-1mm}
\caption{Ablation of component and semantic filtering.}
\label{tab:retrieval_filtering_ablation}
\vspace{-5mm}
\end{table}

\begin{table}[t]
\centering
\scriptsize
\setlength{\tabcolsep}{3.0pt}
\renewcommand{\arraystretch}{1.05}
\begin{tabular*}{0.95\columnwidth}{@{\extracolsep{\fill}}lcc|ccc}
\toprule
\multirow{2}{*}{Retrieval Strategy}
& \multicolumn{2}{c|}{Retrieval Quality (Top-1)}
& \multicolumn{3}{c}{Generation Quality} \\
\cmidrule(lr){2-3} \cmidrule(lr){4-6}
& Text Sim.$\uparrow$ & Gloss Cov.$\uparrow$
& R$\uparrow$ & B4$\uparrow$ & MPJPE$\downarrow$ \\
\midrule
Random retrieval
& 3.83 & 7.39 & 30.70 & 9.68 & 25.45 \\
Sem. similarity
& 44.70 & 39.69 & 32.95 & 10.81 & 24.43 \\
Text overlap
& 54.54 & 45.01 & 34.42 & 11.43 & 22.69 \\
\midrule
Hybrid retrieval
& \textbf{58.87} & \textbf{45.01}
& \textbf{35.09} & \textbf{12.29} & \textbf{22.47} \\
\bottomrule
\end{tabular*}
\vspace{-1mm}
\caption{Analysis of sentence-level retrieval strategies.}
\label{tab:retrieval_strategy}
\vspace{-5mm}
\end{table}

\begin{figure*}[t]
  \centering

    \includegraphics[width=2.0\columnwidth]{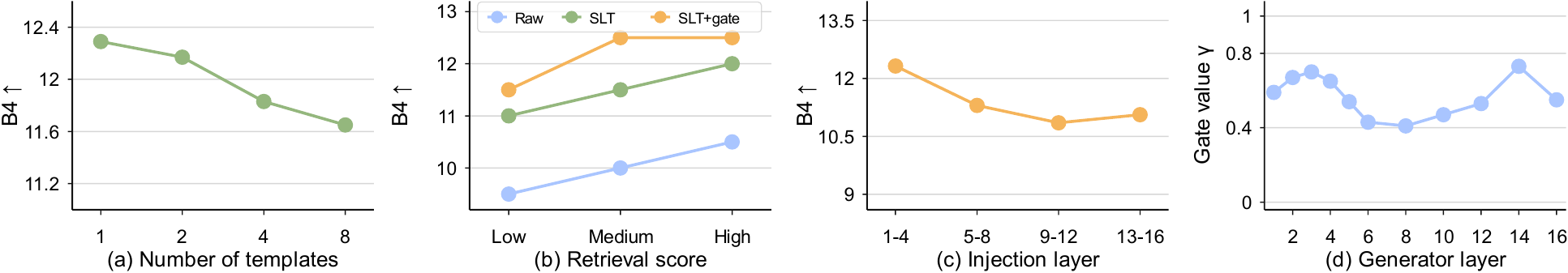}
    \vspace{-3mm}
   \caption{
   Analysis of SemRAG.
(a) Number of retrieved templates.  
(b) Effect of retrieval score. 
(c) Injection layer analysis.
(d) Layer gate visualization.
   }
   \label{fig:gate} 
  \vspace{-5mm}
\end{figure*} 

\begin{figure*}[t]
  \centering
    \includegraphics[width=2.0\columnwidth]{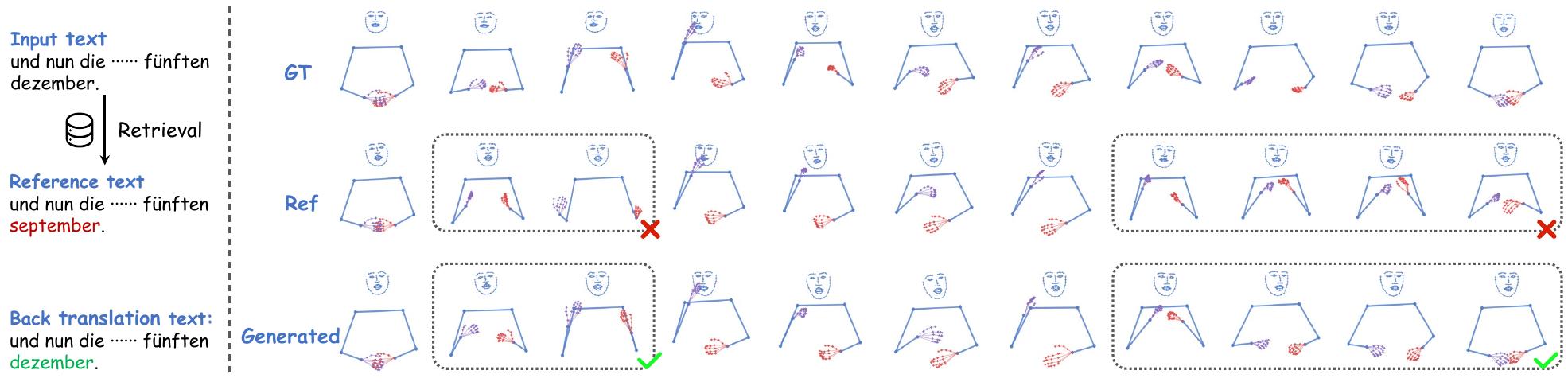}
    \vspace{-3mm}
   \caption{Effect of semantic template filtering.}
   \label{fig:vis} 
  \vspace{-4mm}
\end{figure*} 

\begin{figure}[t]
  \centering
    \includegraphics[width=1.0\columnwidth]{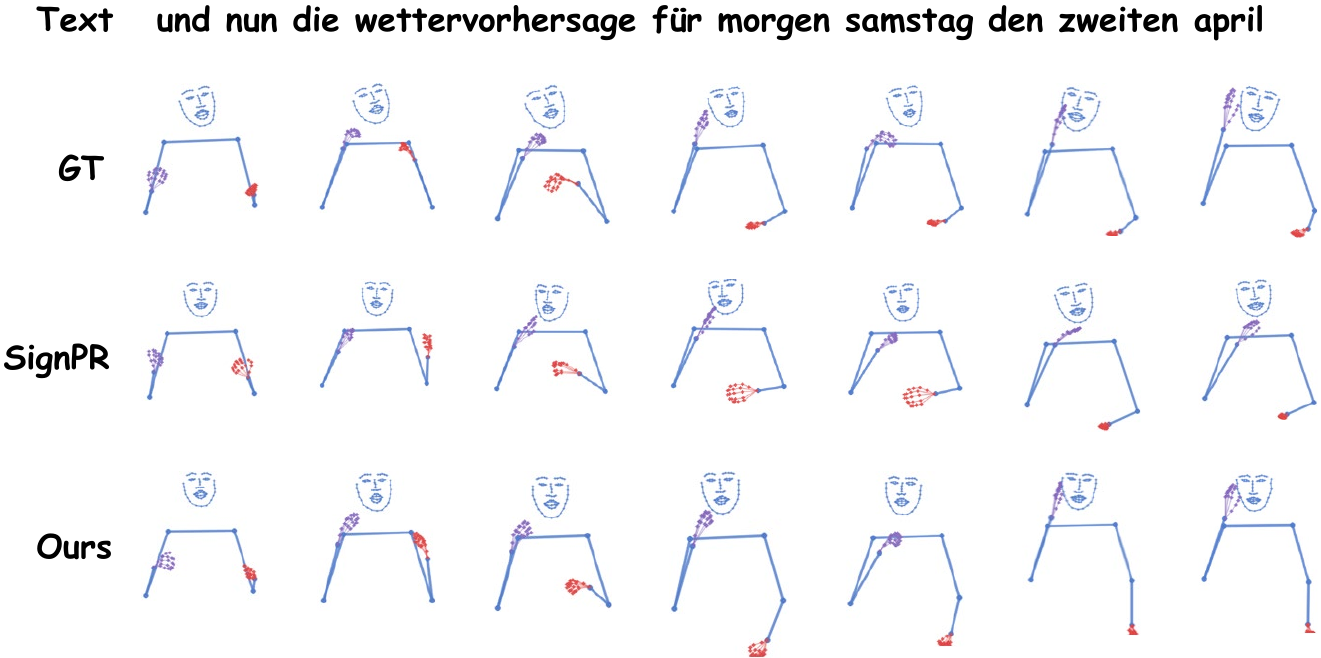}
    \vspace{-5mm}
   \caption{
   Qualitative comparison.
   }
   \label{fig:com} 
  \vspace{-5mm}
\end{figure}

\subsection{Comparison}
Following SignPR, we report ROUGE-L and BLEU-4 using MixSignGraph~\cite{gan2025mixsigngraph} as the back-translation evaluator, with all methods evaluated by the same SLT model for fair comparison.
\textit{We compare SemRAG with text-to-pose and retrieval-based methods, excluding gloss-dictionary-based approaches that require additional gloss annotations.}

\paragraph{Phoenix14T.}
As shown in Table~\ref{tab:SLP}, SemRAG achieves the strongest performance on Phoenix14T. Compared with methods without retrieval, SemRAG improves BLEU-4 over SignPR from 9.41 to 12.29 (+2.88) and reduces MPJPE from 23.04 to 22.47 (-0.57). Compared with retrieval-based baselines, SemRAG also obtains higher back-translation scores than ReMoDiffuse and SOKE. 
Additionally, the small gap between SemRAG$^a$ and SemRAG$^b$ suggests that the filtering mechanism is robust to the choice of SLT model.

\paragraph{CSL-Daily.}
On CSL-Daily, retrieval-based methods generally outperform methods without retrieval, suggesting that external signing references are helpful. SemRAG further improves over retrieval-based baselines: compared with SOKE, it raises BLEU-4 from 10.31 to 12.37 (+2.06) on the test split. Compared with SignPR, SemRAG achieves a large semantic gain, improving BLEU-4 from 3.01 to 12.37 (+9.36). These results further support the use of sentence-level retrieved templates as continuous sign priors for SLP.

\subsection{Ablation Study}
Table~\ref{tab:retrieval_filtering_ablation} 
evaluates the contribution of each component and the effectiveness of SLT-guided filtering.
Adding retrieved sentence templates improves over the text-only baseline, indicating that sentence-level signing examples provide useful motion priors beyond text conditioning.
Applying SLT-guided filtering further improves both semantic and motion metrics, showing that retrieved templates may contain irrelevant motion regions and benefit from selective temporal reweighting.
Finally, layer-gated injection achieves the best performance, demonstrating that adaptive control of retrieval strength helps stabilize template conditioning.

We further compare temporal gate variants for semantic template filtering.
Shuffling the SLT gate disrupts the temporal correspondence between relevance scores and template motions, while inversion emphasizes irrelevant regions.
Both variants degrade performance, confirming that SLT-derived attention provides token-to-time relevance cues for selecting informative regions from retrieved templates.

\subsection{Retrieval and Injection Analysis}

\paragraph{Retrieval strategy.}
Table~\ref{tab:retrieval_strategy} compares sentence-level retrieval strategies.
Random retrieval performs the worst, showing that irrelevant sentence templates provide limited benefit.
Semantic similarity retrieval improves global relevance, while text-overlap retrieval achieves higher gloss coverage and better generation quality, showing the importance of local content matching.
The hybrid retriever combines both semantic and lexical cues and achieves the best performance, which is adopted as the default retrieval strategy.

\paragraph{Number of retrieved templates.}
Figure~\ref{fig:gate}(a) studies the effect of using multiple retrieved templates.
Adding more templates does not improve performance and gradually degrades generation quality, likely because lower-ranked templates introduce additional semantic mismatch.
We therefore use a single retrieved template by default.

\paragraph{Effect of Retrieval Score.}
Figure~\ref{fig:gate}(b) divides test samples into three equal-sized groups based on their top-1 hybrid retrieval scores, labeled Low, Medium, and High.
Generation performance generally improves with higher retrieval scores, while raw template conditioning remains sensitive to mismatched templates. SLT filtering and layer gates consistently improve performance across all groups.

\paragraph{Injection layer analysis.}
Figure~\ref{fig:gate}(c) studies the effect of injecting retrieved template features at different generator depths.
Different injection locations yield varying performance, indicating that retrieved templates provide guidance across generation stages.
We therefore adopt learnable layer gates to regulate retrieval strength across layers.

\paragraph{Layer gate visualization.}
Figure~\ref{fig:gate}(d) visualizes the learned gate values $\gamma_l$ across condition Transformer layers.
The gates vary across layers, indicating that retrieval information is utilized with different strengths during generation.
This supports our layer-gated design, which controls retrieval strength instead of uniformly injecting template features across layers.

\subsection{Human Evaluation}
To evaluate perceptual quality beyond automatic metrics, we conduct a human evaluation with 10 participants on 20 samples per dataset.
The evaluation focuses on the visual quality of generated signing sequences, including motion naturalness and similarity to the ground-truth sequences.
Participants rate each sequence on a five-point scale under identical visualization settings.
More details are provided in the appendix.
As shown in Table~\ref{tab:human_perceptual_study}, SemRAG achieves higher perceptual scores than baselines, indicating improved motion quality and consistency with reference signing sequences.

\begin{table}[t]
\centering
\scriptsize
\setlength{\tabcolsep}{4.2pt}
\renewcommand{\arraystretch}{1.08}

\begin{tabular*}{0.90\columnwidth}{@{\extracolsep{\fill}}lcccc}
\toprule
\multirow{2}{*}{Method}
& \multicolumn{2}{c}{Phoenix14T}
& \multicolumn{2}{c}{CSL-Daily} \\
\cmidrule(lr){2-3}
\cmidrule(lr){4-5}
& Naturalness $\uparrow$
& Motion sim. $\uparrow$
& Naturalness $\uparrow$
& Motion sim. $\uparrow$ \\
\midrule
SignPR
& 3.20
& 3.10
& 3.05
& 2.96 \\

SOKE
& 3.43
& 3.25
& 3.31
& 3.17 \\

\textbf{SemRAG}
& \textbf{3.80}
& \textbf{3.70}
& \textbf{3.92}
& \textbf{3.84} \\
\bottomrule
\end{tabular*}

\vspace{-1mm}
\caption{Human perceptual evaluation on Phoenix14T and CSL-Daily. Participants rate motion naturalness and motion similarity to the ground-truth reference on a five-point scale.}
\label{tab:human_perceptual_study}
\vspace{-5mm}
\end{table}

\subsection{Qualitative Results}

\paragraph{Template usage.}
Figure~\ref{fig:vis} shows a case where the retrieved sentence is similar to the input but differs in key content words. SemRAG uses the retrieved sequence as a reference yet preserves target semantics, as indicated by the back-translation result. This shows the role of SLT filtering and gated injection in suppressing mismatched template content.

\paragraph{Motion quality.}
Figure~\ref{fig:com} compares generated poses from SignPR and SemRAG.
SemRAG produces more coherent motion and better preserves temporal dynamics, especially around continuous transitions.
The qualitative results are consistent with the improvements observed in back-translation metrics and DTW-PA-JPE.
 
\section{Conclusion}

We introduce SemRAG, a semantic-guided sentence-level retrieval-augmented framework for sign language production.
Instead of relying on gloss/word-pose dictionaries or isolated sign segments, SemRAG retrieves full-sentence signing templates as motion priors and uses a frozen SLT model to localize input-relevant regions within retrieved templates.
The resulting template features are then incorporated into the generator through layer-specific learnable gates, which adaptively regulate their contribution across generator layers to improve semantic alignment and motion quality.
Future work could explore larger-scale template collections and extensions to video level generation.

\bibliography{bib/new, bib/newother, bib/newslt}

\clearpage
\newpage
\appendix




\end{document}